%% file: main.tex
\documentclass[10pt,journal,compsoc]{IEEEtran}

\input{packages.tex}
\begin{document}

\title{SelfPose: 3D Egocentric Pose Estimation\\ from a Headset Mounted Camera\\
}

\author{Denis Tome,
        Thiemo Alldieck,
        Patrick Peluse,
        Gerard Pons-Moll\\
        Lourdes Agapito,
        Hernan Badino,
        Fernando de la Torre
        }

\markboth{PAMI Special Issue on Egocentric Perception}%
{}

\IEEEtitleabstractindextext{%
\begin{abstract}
We present a new solution to egocentric 3D body pose estimation from monocular images captured from a downward looking fish-eye camera installed on the rim of a head mounted virtual reality device. This unusual viewpoint leads to images with unique visual appearance, characterized by severe self-occlusions and strong perspective distortions that result in a drastic difference in resolution between lower and upper body. We propose a new encoder-decoder architecture with a novel multi-branch decoder designed specifically to account for the varying uncertainty in 2D joint locations. Our quantitative evaluation, both on synthetic and real-world datasets, shows that our strategy leads to substantial improvements in accuracy over state of the art
egocentric pose estimation approaches. To tackle the severe lack of labelled training data for egocentric 3D pose estimation we also introduced a large-scale photo-realistic synthetic dataset. {\boldmath $x$}R-EgoPose offers $383$K frames of high quality renderings of people with diverse skin tones, body shapes and clothing, in a variety of backgrounds and lighting conditions, performing a range of actions. Our experiments show that the high variability in our new synthetic training corpus leads to good generalization to real world footage and to state of the art results on real world datasets
with ground truth. Moreover, an evaluation on the Human3.6M benchmark shows that the performance of our method is on par with top performing approaches on the more classic problem of 3D human pose from a third person viewpoint.
\end{abstract}

}

\maketitle

\IEEEdisplaynontitleabstractindextext

%
\IEEEpeerreviewmaketitle

\IEEEraisesectionheading{\section{Introduction}\label{sec:introduction}}
\input{sections/introduction.tex}

\section{Related Work}\label{sec:related_work}
\input{sections/related_work.tex}


\section{{\boldmath $x$}R-EgoPose Synthetic Dataset}\label{sec:dataset}
\input{sections/dataset.tex}

\section{Architecture}\label{sec:architecture}
\input{sections/architecture.tex}

\section{Experimental Evaluation}\label{sec:experiments}
\input{sections/experiments.tex}

\section{Conclusion}\label{sec:conclusion}
\input{sections/conclusion.tex}


%



\ifCLASSOPTIONcaptionsoff
  \newpage
\fi


\bibliographystyle{IEEEtran}
\bibliography{egbib}
%
%


\vspace{-15mm}

\begin{IEEEbiography}[{\includegraphics[width=1in,height=1.25in,clip,keepaspectratio]{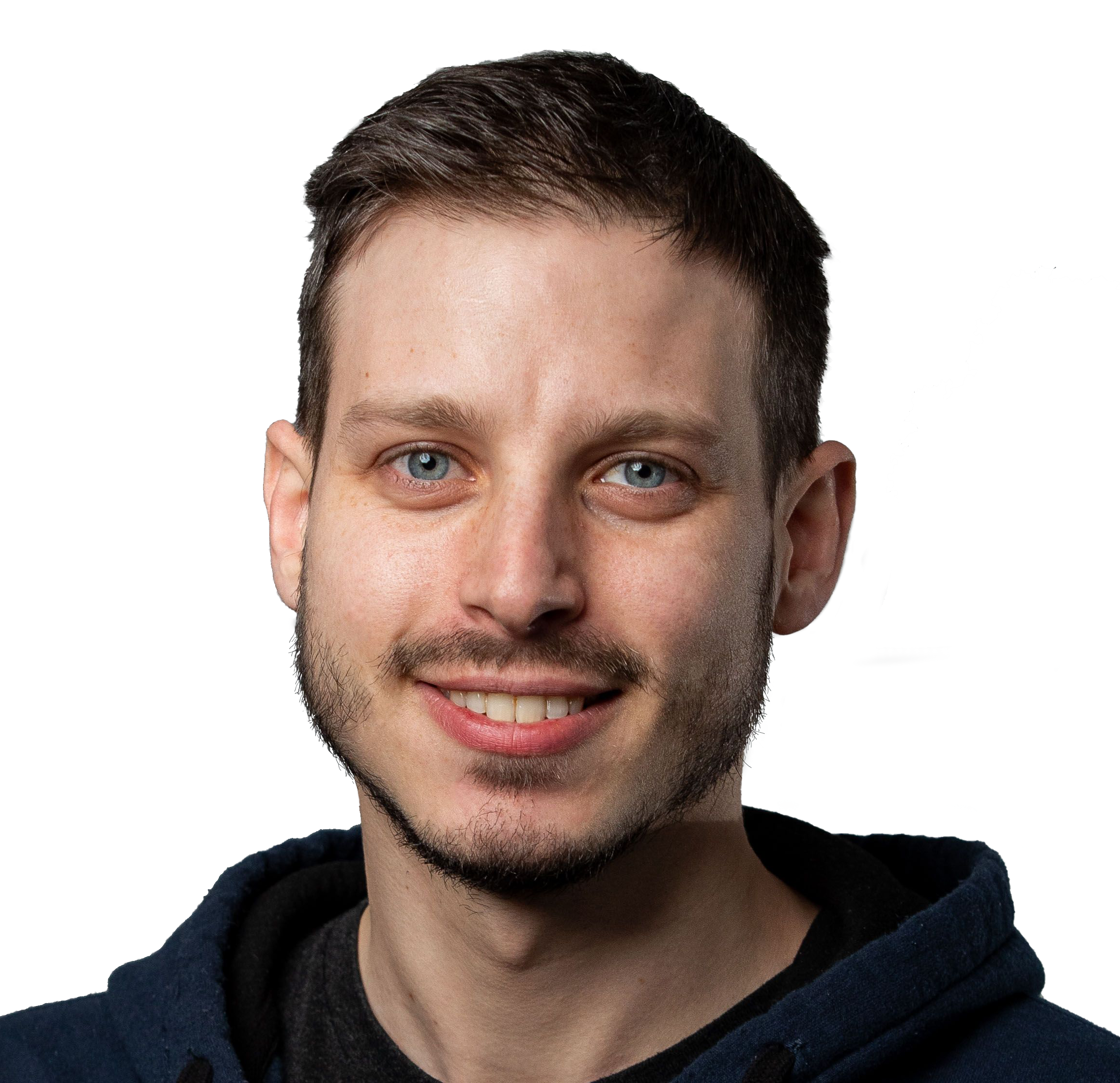}}]{Denis Tome}
is a research scientist at \textit{Epic Games}, working at the intersection between computer vision and graphics. Since 2016 he is a Ph.D.\ candidate and member of the Vision and Imaging Science Group at University College London (UCL) under the supervision of Prof.\ Lourdes Agapito and Dr.\ Gabriel Brostow. His mainly research focuses on 3D human pose reconstruction, both sparse and dense, from simplistic configuration (monocular) to more complex ones (multi-view), for different applications ranging from robotics to VR/AR for headset mounted camera systems. \end{IEEEbiography}

\vspace{-12mm}

\begin{IEEEbiography}[{\includegraphics[width=1in,height=1.25in,clip,keepaspectratio]{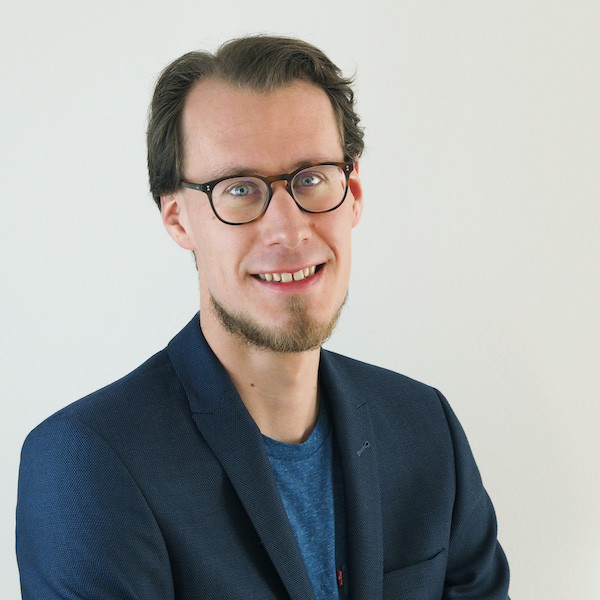}}]{Thiemo Alldieck}
is currently a research intern at \textit{Facebook Reality Labs}. Since 2016 he is a Ph.D.\ candidate at the Computer Graphics Lab at TU Braunschweig, Germany. Since 2018, he is also affiliated with the ``Real Virtual Humans'' group at Max Planck for Informatics (MPII) in Saarbrücken, Germany. His work lies at the intersection between computer vision, graphics, and machine learning and focuses on  human pose, shape, and clothing reconstruction from monocular images and video. \end{IEEEbiography}

\vspace{-12mm}

\begin{IEEEbiography}[{\includegraphics[width=1in,height=1.25in,clip,keepaspectratio]{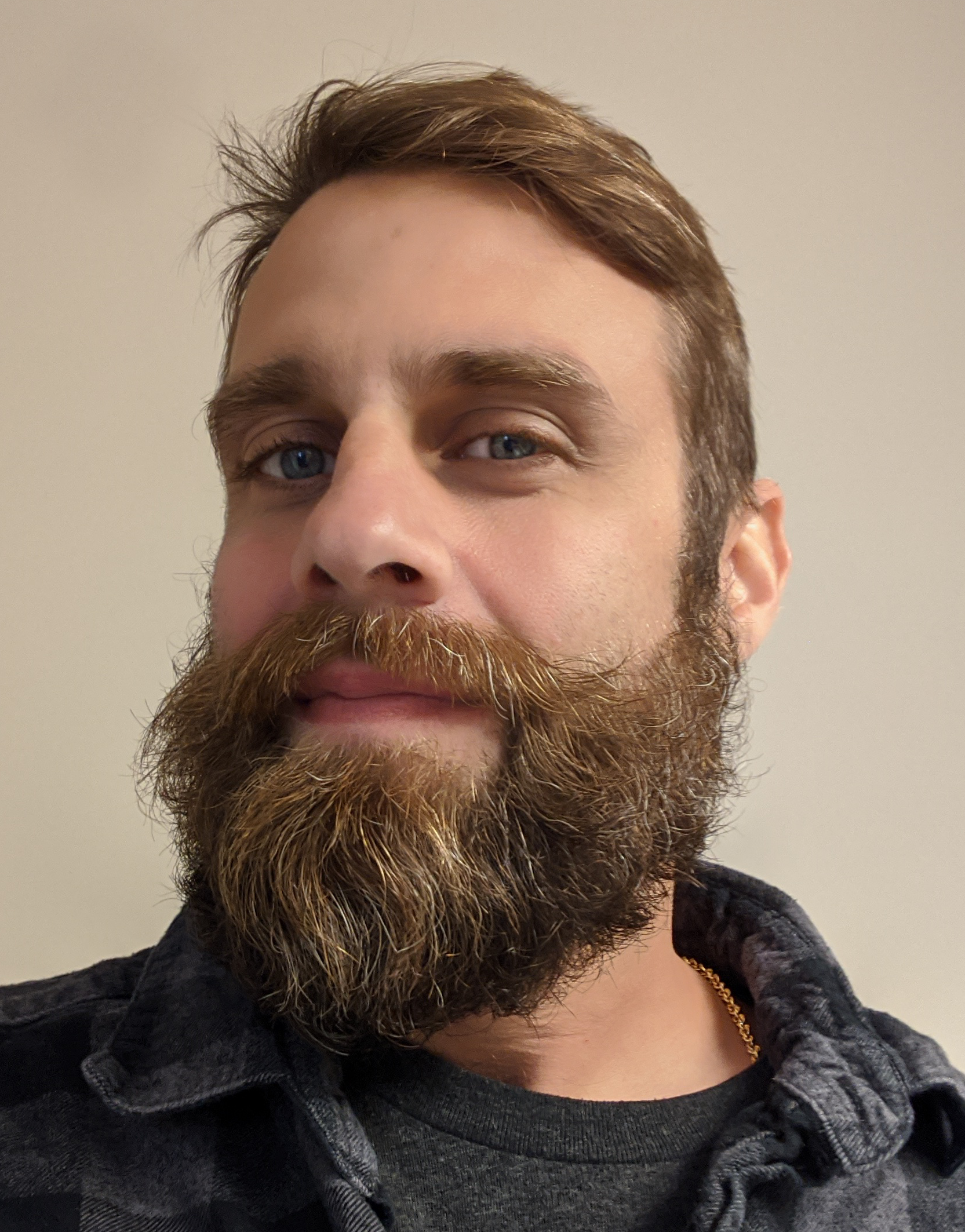}}]{Patrick Peluse}
received the BFA degree in Visual Effects/Animation at the Academy Of Art University in 2008. He has worked in computer graphics for over 15 years with current interests in simulation to solve real world problems. Currently, he is a Technical Artist at Facebook Reality Labs where his focus is on both real-time and offline data generation for quality assurance and hardware prototyping. He has worked as video game programmer, UI/UX conceptual designer for AR/VR, and data generator~/~content creator with companies Autodesk M\&E, Digital Domain, Circuits for Fun, Meta Vision, Magic Leap, and Oculus. \end{IEEEbiography}

\vspace{-12mm}

\begin{IEEEbiography}[{\includegraphics[width=1in,height=1.25in,clip,keepaspectratio]{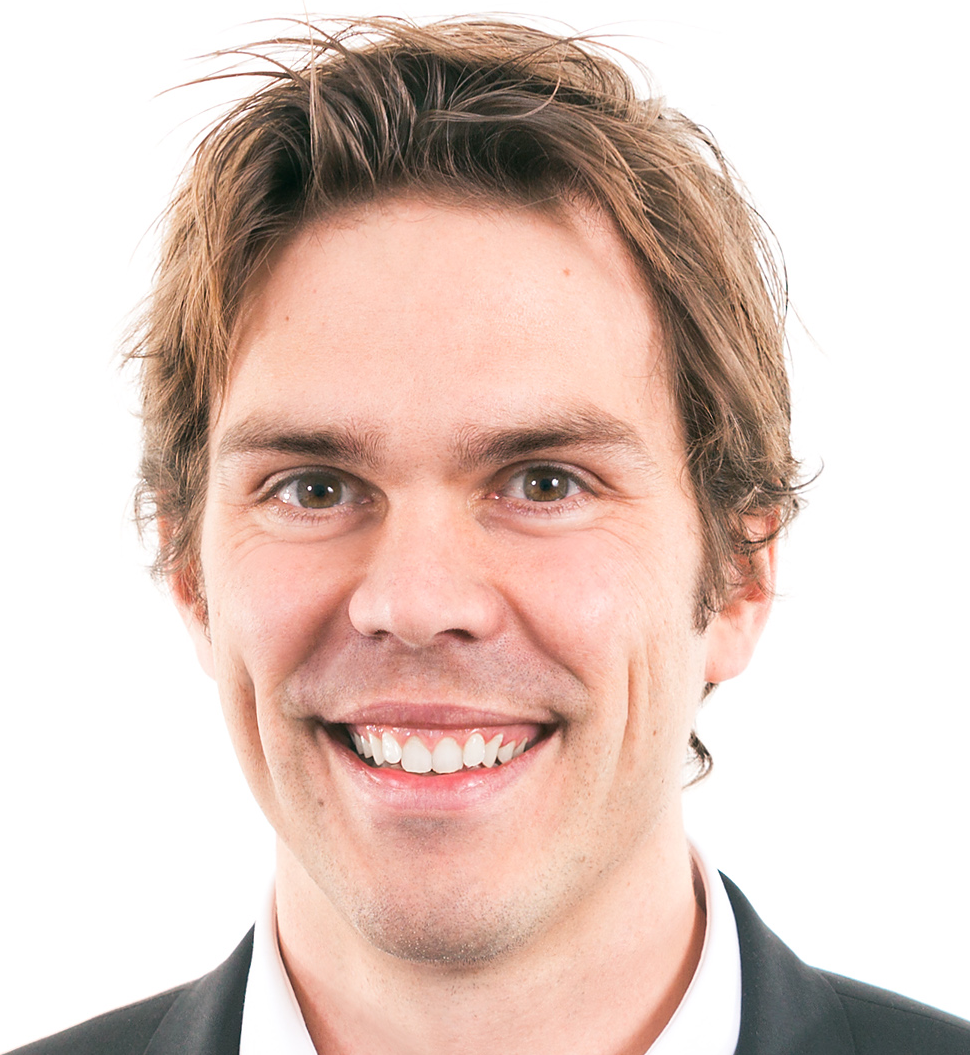}}]{Gerard Pons-Moll}
is the head of the Emmy Noether independent research group ``Real Virtual Humans'', senior researcher at the Max Planck for Informatics (MPII) in Saarbrücken, Germany, and Junior Faculty at Saarland Informatics Campus. His research lies at the intersection of computer vision, computer graphics and machine learning -- with special focus on analyzing people in videos, and creating virtual human models by "looking" at real ones". His work has received several awards including the prestigious Emmy Noether Grant (2018), a Google Faculty Research Award (2019), a Facebook Reality Labs Faculty Award (2018), and recently the German Pattern Recognition Award (2019), which is given annually by the German Pattern Recognition Society to one outstanding researcher in the fields of Computer Vision and Machine Learning. His work got Best Papers Awards at BMVC’13, Eurographics’17 and 3DV’18 and he served as Area Chair for ECCV'18, 3DV'19, SCA'18'19, FG'20 and will serve as Area Chair for CVPR'21, ECCV'20 and 3DV'20. \end{IEEEbiography}

\vspace{-12mm}

\begin{IEEEbiography}[{\includegraphics[width=1in,height=1.25in,clip,keepaspectratio]{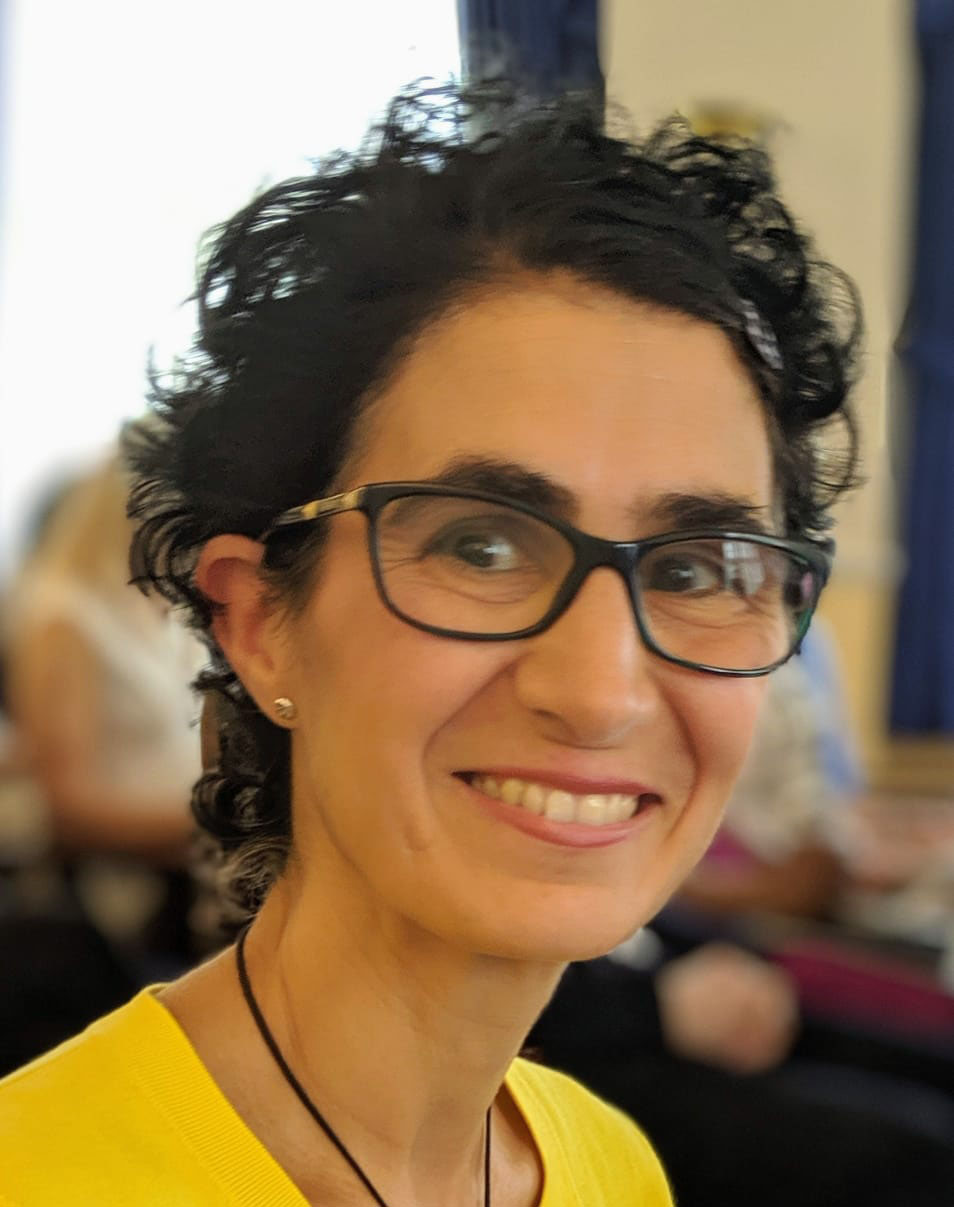}}]{Lourdes Agapito}
is Professor of 3D Vision and Head of the Vision and Imaging Science Group in the Department of Computer Science at  University College London (UCL). She is also co-founder of London-based startup Synthesia Technologies and was an ERC Grant holder (2008-14). She obtained her BSc, MSc and PhD from the Universidad Complutense (Madrid) in 1991, 1992 and 1996, and was a postdoctoral fellow at the University of Oxford's Robotics Research Group (1997-2001). Her research interests lie at the intersection of computer vision, graphics and machine learning; more specifically 3D reconstruction from video, 3D shape modelling, weakly supervised learning for 3D vision, human pose estimation and video synthesis. She has served as Program Chair for the top computer vision conferences (CVPR'16, ICCV'21), Workshops Chair for ECCV'14 and Area chair for CVPR (3x), ECCV (2x), ICCV (1x). She is associate editor of IEEE PAMI and IJCV. She was keynote speaker at ICRA'17. \end{IEEEbiography}

\vspace{-12mm}

\begin{IEEEbiography}[{\includegraphics[width=1in,height=1.25in,clip,keepaspectratio]{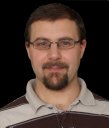}}]{Hernan Badino}
is a research scientist at Facebook Reality Labs. He received his PhD degree in Computer Sciences from the Goethe University Frankfurt in 2008. During his PhD, he worked with the Image Based Environment Perception Group at Daimler AG and joined the Carnegie Mellon University in 2009 as a postdoctoral fellow. In 2012, he was appointed a faculty position at the Robotics Institute at the Carnegie Mellon University where he worked on visual based localization, sensor-fusion, ego-pose estimation, object detection, and tracking, and real-time embedded solutions for visual-based pose estimation. He joined Facebook Reality Labs in 2015 and has since then been working at the intersection of computer vision, motion capture, and telepresence systems towards the goal of achieving social presence in artificial reality.\end{IEEEbiography}

\vspace{-12mm}

\begin{IEEEbiography}[{\includegraphics[width=1in,height=1.25in,clip,keepaspectratio]{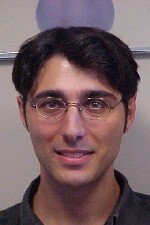}}]{Fernando de la Torre}
received the BSc degree in telecommunications, and the MSc and PhD degrees in electronic engineering from the La Salle School of Engineering at Ramon Llull University, Barcelona, Spain in 1994, 1996, and 2002, respectively. He is an associate research professor in the Robotics Institute at Carnegie Mellon University. His research interests are in the fields of computer vision and machine learning. Currently, he is directing the Component Analysis Laboratory (http://ca.cs.cmu.edu) and the Human Sensing Laboratory (http://humansensing.cs.cmu.edu) at Carnegie Mellon University. He has more than 100 publications in refereed journals and conferences. He has organized and co-organized several workshops and has given tutorials at international conferences on the use and extensions of component analysis.
\end{IEEEbiography}

\end{document}

%% file: packages.tex
\ifCLASSOPTIONcompsoc
  \usepackage[nocompress]{cite}
\else
  \usepackage{cite}
\fi
\usepackage{xspace}

\ifCLASSINFOpdf
  \usepackage[pdftex]{graphicx}
  \graphicspath{figures/}
  \DeclareGraphicsExtensions{.pdf,.jpeg,.png}
\else
   \usepackage[dvips]{graphicx}
  \graphicspath{figures/}
  \DeclareGraphicsExtensions{.pdf,.jpeg,.png}
\fi

\usepackage{xcolor}

\usepackage{amsmath}
\usepackage{amssymb}

\usepackage{array}
\usepackage{multirow}
\usepackage{booktabs}

\usepackage{caption, subcaption}

\usepackage{hyperref}

%% file: sections/introduction.tex
%

\IEEEPARstart{T}{he} advent of {\boldmath $x$}R technologies (such as AR, VR, and MR) has led to a wide variety of applications in areas such as entertainment, communication, medicine, CAD design, art, and workspace productivity. These technologies mainly focus on immersing the user in a virtual space using a head mounted display (HMD) which renders the environment from the specific viewpoint of the user. However, current solutions have  so far focused on the video and audio aspects of the user's perceptual system, leaving a gap in the touch and proprioception senses. Partial solutions to  proprioception have been limited to the use of controller devices to track and render hand  positions in real time. The 3D pose of the rest of the body is then inferred from inverse kinematics of the head and hand poses \cite{invkinavatars18}, but this often results in inaccurate estimates of the body configuration with a large loss of signal that impedes compelling social interaction \cite{uhess16} and even leads to motion sickness \cite{reason1975motion}.

\hfill

\begin{figure*}[tb]
  \includegraphics[width=\linewidth]{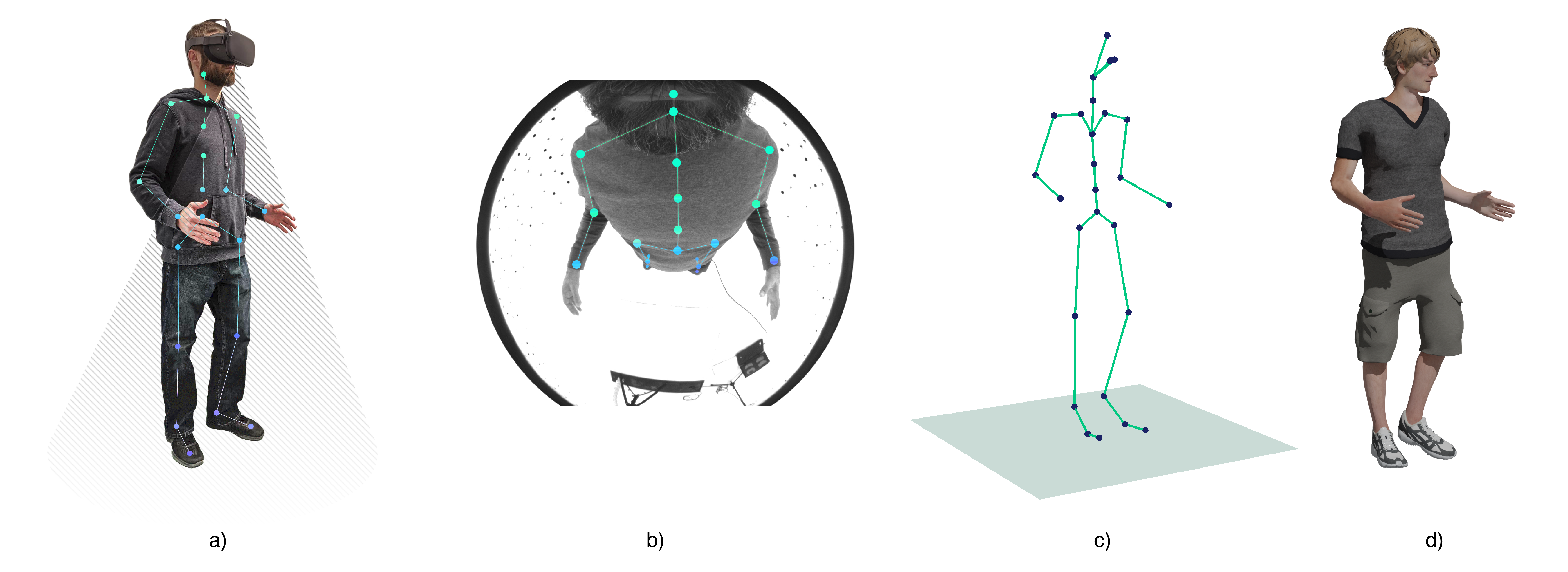}
  \caption{Egocentric human pose estimation: driving an avatar from an egocentric camera perspective.
  \emph{b)} Egocentric perspective of the pose visualized in \emph{a)} from an external point of view; \emph{c)} 3D joint locations predicted from the input RGB only-information shown in \emph{b)}; \emph{d)} synthetic character driven from the local joint rotations estimated alonside the 3D locations. \label{fig:intro:pipeline}}
\end{figure*}

Fig.~\ref{fig:intro:pipeline} illustrates the problem that this paper addresses: the goal is to infer 2D and 3D pose information, such as joint positions and rotations, from an egocentric camera perspective, necessary to transfer the motion from the original user to a \emph{generic avatar} or to gather user pose information. 

\begin{figure*}[t]
\includegraphics[width=\linewidth]{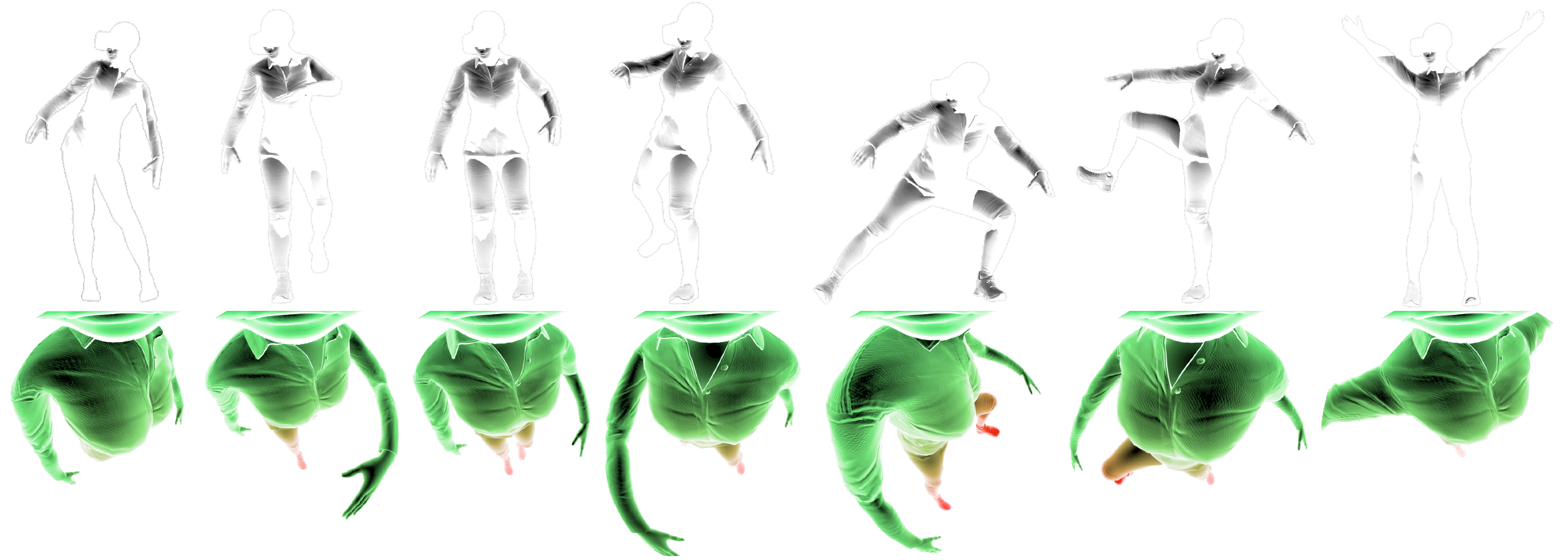}
\caption{Visualization of different poses with the same character. {\bf Top:} poses rendered from an external camera
viewpoint. White represents occlusion, which is body parts that would not be visible from the egocentric perspective.
{\bf Bottom:} poses rendered from the egocentric camera viewpoint. Color gradient indicates the density of image pixels
for each area of the body: \textit{green} is higher pixel density, whereas \textit{red} is lower density. This figure
illustrates the challenges faced in egocentric human pose estimation: severe self-occlusions, extreme perspective effects
and lower pixel density for the lower body. \label{fig:intro:perspective}}
\end{figure*}

The monocular camera used in our configuration is mounted on the rim of a HMD (as shown in Fig.~\ref{fig:intro:pipeline}a), approximately $2cm$ away from an average size nose, looking down. Fig.~\ref{fig:intro:perspective} provides a more clear visualization of the unique visual appearance of the images that the camera sees for different body configurations --- the top row shows which body parts would become self-occluded from an egocentric viewpoint. The continuous gradation from bright red to dark green encodes the increasing pixel resolution for the corresponding colored area.

There are several challenges that contribute to the difficulty of this problem: \emph{(1)} Strong perspective distortions occur, due to the fish-eye lenses and the proximity of the camera to the face. This results in images with strong radial distortion and drastic difference in image resolution between the upper and lower body (as visible in Fig.~\ref{fig:intro:perspective} --- bottom row). Due to this, state-of-the-art approaches for 2D body pose estimation~\cite{rogez2019lcr} from a frontal or 360 degree yaw view, fail on this type of images; \emph{(2)} There are many instances where body self-occlusion occurs, especially in the lower-body (see right images of Fig.~\ref{fig:comparison_mo2cap2}), which demands strong spatial awareness of joint locations; \emph{(3)} Egocentric 3D body pose estimation is a relatively unexplored problem in computer vision, hence the scarce availability of publicly accessible labeled datasets; \emph{(4)} As shared by traditional 3D body pose estimation, natural ambiguity is present when lifting 2D joint positions in 3D.

\hfill

The unusual visual egocentric appearance calls for a new approach and a new training corpus. This paper tackles both. Our novel neural network architecture encodes the difference in uncertainty between upper and lower body joints caused by the varying resolution, extreme perspective effects and
self-occlusions.

We conducted quantitative and qualitative evaluations on both synthetic and real-world benchmarks with ground truth 3D annotations, showing that our approach outperforms previous egocentric state-of-the-art $\text{Mo}^2\text{Cap}^2$~\cite{xu2019mo2cap2} by \textbf{more than 25\%}. In addition, we achieve state-of-the-art performance on the more standard front-facing cameras 3D human pose reconstruction scenario, without any architecture modifications, performing second best  after~\cite{sun2018integral} on the Human3.6M benchmark~\cite{ionescu2014human3}.

Our ablation studies show that the introduction of our novel \emph{multi-branch} decoder to reconstruct the 2D input heatmaps and rotations, is responsible for the drastic improvements in 3D pose estimation. Furthermore, the contribution of each of the branches is analyzed, providing tools to control the level of uncertainty embedded in the latent space.

%% file: sections/related_work.tex
We describe related work on monocular (single-camera) marker-less 3D human pose estimation focusing on two distinct capture setups: \emph{outside-in} approaches where an external camera viewpoint is used to capture one or more subjects from a distance -- the most commonly used setup; and \emph{first person} or egocentric systems where a head-mounted camera observes the own body of the user. While our paper focuses on the second scenario, we build on recent advances in CNN-based methods for human 3D pose estimation. We also describe approaches that incorporate wearable sensors for first person human pose estimation.

\hfill

\noindent{\bf Monocular 3D Pose Estimation from an External Camera Viewpoint:}
the advent of convolutional neural networks and the availability of large 2D and 3D training datasets~\cite{ionescu2014human3,andriluka142d} has recently allowed fast progress in monocular 3D pose estimation from RGB images captured from external cameras. Two main trends have emerged: \emph{(i)} fully supervised regression of 3D joint locations directly from images ~\cite{li20143D,park20163D,tekin2016structured,zhou2016deep,pavlakos2017coarse,mehta2017monocular} and \emph{(ii)} pipeline approaches that decouple the problem into the tasks of 2D joint detection followed by 3D lifting~\cite{martinez2017simple,moreno20173D,ramakrishna2012reconstructing,akhter2015pose,zhou2017sparse,zhou2016sparseness,bogo2016keep,sanzari2016bayesian,pavlakos2019expressive,alldieck2018video}. Progress in fully supervised approaches and their ability to generalize has been severely affected by the limited availability of 3D pose annotations for in-the-wild images. This has led to significant efforts in creating photo-realistic synthetic datasets~\cite{rogez2016mocap,varol2017learning} aided by the recent availability of parametric dense 3D models of the human body learned from body scans~\cite{loper2015smpl}. On the other hand, the appeal of
two-step decoupled approaches comes from two main advantages: the availability of high-quality off-the-shelf 2D joint detectors~\cite{wei2016convolutional,newell2016stacked,pishchulin2016deepcut,cao2017realtime} that only require easy-to-harvest 2D annotations, and the possibility of training the 3D lifting step using 3D mocap datasets and their ground truth projections without the need for 3D annotations for images. Even simple architectures have been shown to solve this task with a low error rate~\cite{martinez2017simple}. Recent advances are due to combining the 2D and 3D tasks into a joint estimation~\cite{rogez2017lcr,rogez2019lcr,mehta2019xnect,mehta2018multiperson} and using weakly~\cite{wu2016single,tome2017lifting,tung2017adversarial,drover2018can,pavlakos2018learning} or self-supervised losses~\cite{tung2017self,rhodin2018unsupervised,omran2018NBF,humanMotionKanazawa19,hmrKanazawa17} or mixing 2D and 3D data for training~\cite{sun2018integral,omran2018NBF,alldieck19cvpr,bhatnagar2019mgn}.

\hfill

\noindent{\bf First Person 3D Human Pose Estimation:} while capturing users from an egocentric camera perspective for activity recognition has received significant attention in recent years~\cite{fathi2011understanding,ma2016going,cao2017egocentric}, most methods detect, at most, only upper body motion (hands, arms or torso). Capturing full 3D body motion from head-mounted cameras is considerably more challenging. Some head-mounted capture systems are based on RGB-D input and reconstruct mostly hand, arm and torso motions~\cite{yonemoto2015egocentric, rogez2015first}. Jiang and Grauman~\cite{jiang2017seeing} reconstruct full body pose from footage taken from a camera worn on the chest by estimating egomotion from the observed scene, but their estimates lack accuracy and have high uncertainty. Yuan~\etal~\cite{yuan20183d,Yuan_2019_ICCV} instead explores a different solution by moving away from kinematics-based representations and using a control-based representation of humanoid motion, commonly used in robotics. A step towards dealing with large parts of the body not being observable was proposed in~\cite{amer2018deep} but for external camera viewpoints. Rhodin \etal~\cite{rhodin2016egocap} pioneered the first approach towards full-body capture from a helmet-mounted stereo fish-eye camera pair. The cameras were placed around 25 cm away from the user's head, using telescopic sticks, which resulted in a fairly cumbersome setup for the user but with the benefit of capturing large field of view images where most of the body was in view.  A monocular head-mounted systems for full-body pose estimation has more recently been demonstrated by Xu~\etal~\cite{xu2019mo2cap2}, who propose a real-time compact setup mounted on a baseball cap, although in this case the egocentric camera is placed a few centimeters further from the user's forehead than in our proposed approach. Our approach substantially outperforms Xu~\etal's method~\cite{xu2019mo2cap2} by at least $20\%$ on both indoor and outdoor sequences from their real world evaluation dataset.
In this journal paper, we go beyond our previous conference paper~\cite{tome2019xr}. First, we perform an extensive analysis on deep architectures for the task of egocentric pose estimation, and show that UNet architectures significantly outperform the originally proposed ResNet architecture~\cite{tome2019xr}, specifically for transfer learning from synthetic to real data. Second, we propose a new model which additionally predicts per part rotations. In contrast to ~\cite{tome2019xr}, this allows us to animate virtual characters, which is necessary for many applications.  

\hfill

\begin{figure*}[t]
  \centering
  \includegraphics[width=\linewidth]{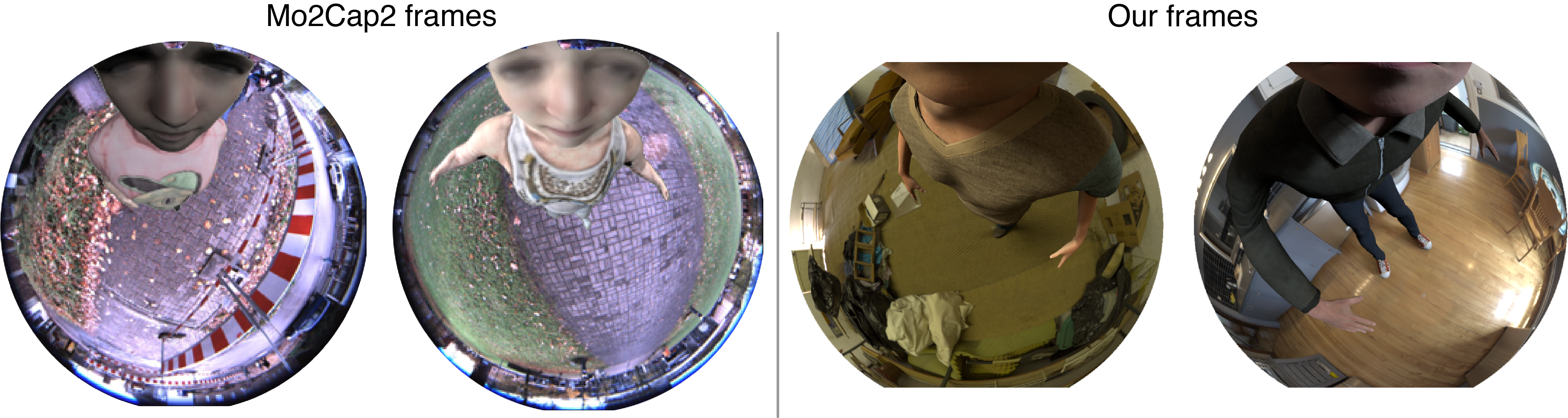}
  \caption{Example images from our {\boldmath $x$}R-EgoPose Dataset
    compared with the competitor Mo2Cap2
    dataset~\cite{xu2019mo2cap2}. The quality of our frames is far
    superior than the randomly sampled frames from mo2cap2, where the
    characters suffer color matching with respect to the background
    light conditions. \label{fig:comparison_mo2cap2}}
\end{figure*}

\noindent{\bf 3D Pose Estimation from Wearable Devices:} 
Inertial Measurement Units (IMUs) worn by the subject provide a camera-free alternative solution to first person human pose estimation. However, such systems are intrusive and complex to calibrate. While reducing the number of sensors leads to a less invasive configuration~\cite{von2017sparse,DIP:SIGGRAPHAsia:2018} recovering accurate human pose from sparse sensor readings becomes a more challenging task. 
Video data can be fused with IMU~\cite{malleson2019real,vonmarcardponsmollPAMI16,ponsmollICCV2011,vonMarcard2018} to improve accuracy, but these approaches require line of sight with an external camera. 
An alternative approach, introduced by Shiratori \etal~\cite{shiratori2011motion} consists of a multi-camera structure-from-motion (SFM) approach using 16 limb-mounted cameras. Still very intrusive, this approach suffers from motion blur, automatic white balancing, rolling shutter effects and motion in the scene, making it impractical in realistic scenarios.

%% file: sections/dataset.tex

Ego-3D posed estimation from HMC is a relatively new research problem in computer vision, and to the best of our knowledge there is only one dataset available to analyze the algorithms, see Fig.~\ref{fig:comparison_mo2cap2}. Existing databases are not rich enough to provide statistical significant analysis due to the scarcity of data. This section describes a photo-realistic synthetic  egocentric dataset with ground-truth data, that overcomes some of the limitations of existing approaches. 

The design of this dataset focuses on scalability, with augmentation of characters, environments, and lighting conditions. A rendered scene is generated from a random selection of characters, environments, lighting rigs, and animation actions. The animations are obtained from mocap data. A small random displacement is added to the positioning of the camera on the headset to simulate the typical variation of the pose of the headset with respect to the head when worn by the user.

\noindent\textbf{Characters}: To improve the diversity of body types,
from a single character, we generate additional \textit{skinny short},
\textit{skinny tall}, \textit{full short}, and \textit{full tall}
versions. The height distribution of ranges from
$155$ cm to $189$ cm.

\noindent\textbf{Skin}: color tones include \textit{white} (Caucasian,
freckles or Albino), \textit{light-skinned European},
\textit{dark-skinned European} (darker Caucasian, European mix),
\textit{Mediterranean or olive} (Mediterranean, Asian, Hispanic,
Native American), \textit{dark brown} (Afro-American, Middle Eastern),
and \textit{black} (Afro-American, African, Middle
Eastern). Additionally, we built random skin tone parameters into the
shaders of each character used with the scene generator.

\noindent\textbf{Clothing}: Clothing types include athletic pants,
jeans, shorts, dress pants, skirts, jackets, T-Shirts, long sleeves,
and tank tops. Shoes include sandals, boots, dress shoes, athletic
shoes, crocs. Each type is rendered with different texture and
colors. 

 \noindent\textbf{Actions}: the type of actions are listed in
 Table~\ref{tab:action_frames}.


\noindent\textbf{Images}: the images have a resolution of $1024 \times
1024$ pixels and 16-bit color depth. For training and testing, we
downsample the color depth to 8 bit. The frame rate is $30$
fps. \textit{RGB}, \textit{depth}, \textit{normals}, \textit{body
  segmentation}, and \textit{pixel world position} images are
generated for each frame, with the option for exposure control for
augmentation of lighting. Metadata is provided for each frame
including 3D joint positions, height of the character, environment,
camera pose, body segmentation, and animation rig.
 

\noindent\textbf{Render quality}: Maximizing the photo-realism of the
synthetic dataset was our top priority. Therefore, we animated the
characters in Maya using actual mocap data \cite{Mixamo19}, and used a
standardized physically based rendering setup with V-Ray. The
characters were created with global custom shader settings applied
across clothing, skin, and lighting of environments for all rendered
scenes.

\subsection{Training, Test, and Validation Sets}
The dataset has a total size of $383$K frames, with $23$ male and $23$
female characters, divided into three sets: \emph{Train-set:} $252$K
frames; \emph{Test-set:} $115$K frames; and \emph{Validation-set:}
$16$K frames. The gender distribution is: \emph{Train-set:} 13M/11F,
\emph{Test-set:} 7M/5F and \emph{Validation-set:}
3M/3F. Table~\ref{tab:action_frames} provides a detailed description
of the partitioning of the dataset according to the different actions.
\begin{table}[htbp]
  \begin{center}
    \small
    \renewcommand{\arraystretch}{1.2}
    \begin{tabular}{lccc}
      \textbf{Action}  & \textbf{N. Frames} & \textbf{Size Train} & \textbf{Size Test} \\
      \toprule
      Gaming           & 24019              & 11153               & 4684               \\
      Gesticulating    & 21411              & 9866                & 4206               \\
      Greeting         & 8966               & 4188                & 1739               \\
      Lower Stretching & 82541              & 66165               & 43491              \\
      Patting          & 9615               & 4404                & 1898               \\
      Reacting         & 26629              & 12599               & 5104               \\
      Talking          & 13685              & 6215                & 2723               \\
      Upper Stretching & 162193             & 114446              & 46468              \\
      Walking          & 34989              & 24603               & 9971               \\
      \bottomrule
    \end{tabular}
  \end{center}
  \caption{Total number of frames per action and their distribution
    between train and test sets. Everything else not mentioned is
    validation data.\label{tab:action_frames}}
\end{table}

\begin{figure*}[t]
  \includegraphics[width=\linewidth]{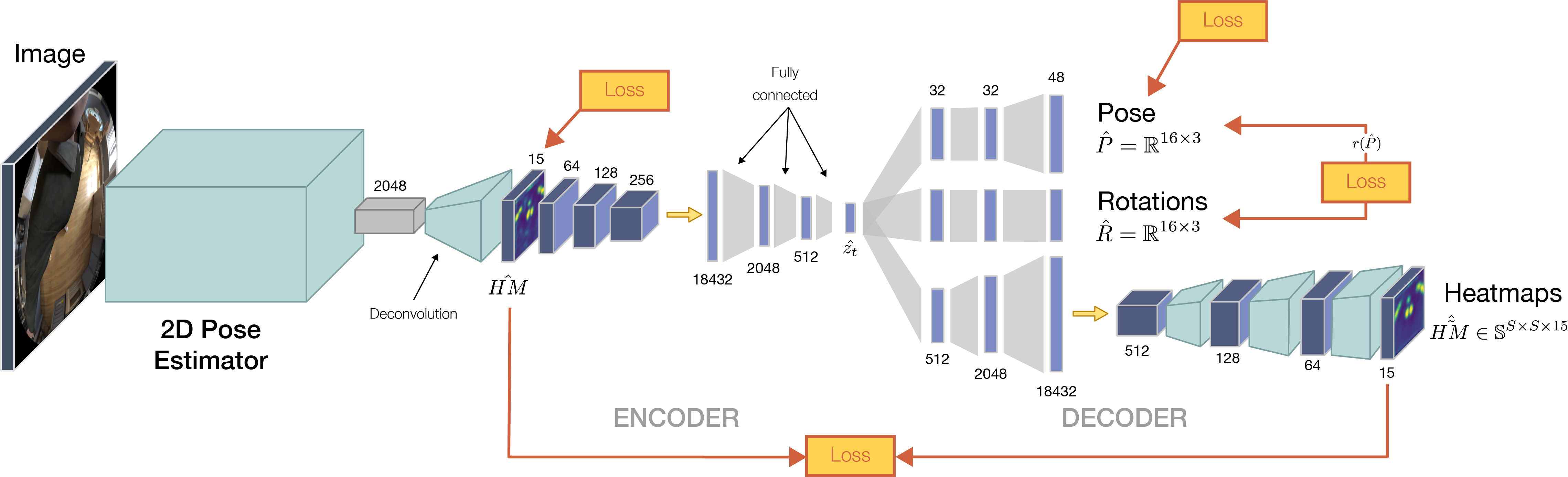}
  \caption{Proposed architecture for egocentric 3D human pose estimation consisting of two modules: \emph{a)} interchangeable 2D pose detector that predicts heatmaps from the input RGB image; \emph{b)} multi-branch auto-encoder that finds a representation of poses which includes also a level of uncertainty of predictions per joint. Alongside the main branch, for 3D joint location prediction, two auxiliary branches as used at training-time to improve latent space distribution. Branch \emph{ii)} estimates local joint rotations, forcing them to be consistent with those rotations extracted by the predicted pose from \emph{i)}; branch \emph{iii)} forces the latent space to include a level of uncertainty of the 2D joint locations by reconstructing the given predicted heatmaps from the pose embedding. These additional branches have demonstrated considerable improvements with respect to a standard AE architecture, as shown in Sec.~\ref{sec:experiments}. \label{fig:architecture}}
\end{figure*}

%% file: sections/architecture.tex
This section describes the deep learning architecture for 3D pose estimation. The proposed architecture (Fig.~\ref{fig:architecture}), is a two step approach consisting of two main modules: \emph{i)} the first module detects 2D heatmaps of the locations of the body joints in image space. We experiment with different standard architectures, please refer to Sec.~\ref{sec:experiments} for details; \emph{ii)}~the second one takes as input the 2D heatmap predictions generated from the preceding module and regresses the 3D coordinates  of the body joints, local joint rotations according to the skeleton hierarchy and reconstructed heatmap predictions, using a novel \emph{multi-branch auto-encoder} architecture.

One of the most important advantages of this pipeline approach is that 2D and 3D modules can be trained independently according to the available training data. For instance, if a sufficiently large corpus of images with 3D annotations is not available, the 3D lifting module can be trained independently using 3D mocap data and its projected heatmaps.  Once the two modules are pretrained the entire architecture can be fine-tuned end-to-end since it is fully differentiable. The \emph{multi-branch} auto-encoder module gives also the ability of having multiple representations of the pose: e.g. joint positions, local rotations, etc. A further advantage of this architecture is that the second and third branches are only needed at training time (see Sec.~\ref{sec:auto_encoder}) and can be removed at test time, guaranteeing the better performance and a faster execution. 

\subsection{2D Pose Detection}
\label{sec:hm_regressor}
Given an RGB image $\m{I} \in \mathbb{R}^{368 \times 368 \times 3}$ as input, the 2D pose detector infers 2D poses, represented as a set
of heatmaps $\m{HM} \in \mathbb{R}^{47 \times 47 \times 15}$, one for each of the body joints.
For this task we have experimented with different standard architectures including \textit{ResNet 50}~\cite{he2016deep} and \textit{U-Net} \cite{ronneberger2015u}. For a detailed analysis, please refer to Sec.~\ref{sec:experiments}.

The models were trained using normalized input images, obtained by subtracting the mean value and dividing by the standard deviation, and using the mean
square error of the difference between the ground truth heatmaps and the predicted ones as the loss:
\begin{equation}
  \text{L}_\text{2D} = \text{mse}(\m{HM}, \widehat{\m{HM}})
\end{equation}

\subsection{2D-to-3D Mapping}
\label{sec:auto_encoder}
The 3D pose module takes as input the $15$ heatmaps computed by the first module and outputs the final 3D pose ${\m{P} \in \mathbb{R}^{16\times3}}$ as a set of joint locations. Note that the number of output 3D joints is $16$ since we include the head which despite being out of the field of view it can be regressed in 3D.

In most pipeline approaches the \textit{3D lifting module} usually is given as input the 2D joint pixel positions in the image of the detected the 3D position. Instead, similarly to Pavlakos~\etal~\cite{pavlakos2018learning}, our approach predicts the 3D pose from input heatmaps, not just 2D locations. The main advantage is that these heatmaps carry important information relative to the \textit{uncertainty of the 2D} pose estimations. Furthermore, due to the unique architecture, it is possible to change the different levels or representation of a pose, afterwards. 

The main novelty of the proposed architecture (see Fig.~\ref{fig:architecture}), is that we ensure that the uncertainty information expressed in the heatmap representations does not get lost but it is preserved in the pose embedding. While the encoder takes as input a set of heatmaps and encodes them into the embedding $\hat{\m{z}}$, the decoder has multiple branches -- \emph{1st} regresses the 3D pose from $\hat{\m{z}}$; \emph{2nd} estimates the local joint rotations (with respect to the parent node); and \emph{3rd} reconstructs the input heatmaps. The purpose of this branch is to force the latent vector to encode the probability density function of the estimated 2D heatmaps.

The overall loss function for the auto-encoder is expressed as
\begin{eqnarray} 
  \text{L}_\text{AE} & = & \lambda_p(||\m{P} - \hat{\m{P}}||^2 + W(\m{P}, \hat{\m{P}})) + \nonumber \\   
  & & \lambda_{r}||\hat{\m{R}} - r({\hat{\m{P}}})||^2 + \nonumber \\
  & & \lambda_{hm}||\widehat{\m{HM}} - \widetilde{\m{HM}}||^2 \label{eq:decoder}
 \end{eqnarray}
with $\m{P}$ the ground truth; $\hat{\m{R}}$ the predicted local joint rotations and $r(\hat{\m{P}})$ the function that estimates local joint rotations from a given pose; $\widetilde{\m{HM}}$ is the set of heatmaps regressed by the decoder from the latent space and $\widehat{\m{HM}}$ are the heatmaps regressed by 2D pose estimator module (see Sec.~\ref{sec:hm_regressor}). Different local joint rotation representations were tested and ultimately a Quaternion representation was chosen due to the stability of the rotations during training, leading to more robust models. The rotation branch also helps generating better results as shown in Sec.~\ref{sec:experiments} with smoother transitions on consecutive frames on poses estimated frame-by-frame.

Finally $W$ is the regularizer over the 3D poses
\begin{align*}
W(\m{P}, \hat{\m{P}}) = \lambda_\theta\theta(\m{P}, \hat{\m{P}}) + \lambda_L\text{L}(\m{P}, \hat{\m{P}})
\end{align*}
with
\begin{align*}
  & \theta(\m{P}, \hat{\m{P}}) = \sum_l^L\frac{\m{P}_l \cdot \hat{\m{P}}_l }{||\m{P}|| * ||\hat{\m{P}}_l||}
  & \text{L}(\m{P}, \hat{\m{P}}) = \sum_l^L||\m{P}_l-\hat{\m{P}}_l||
\end{align*}
corresponding to the cosine-similarity error and the limb-length error, with $\m{P}_l \in \mathbb{R}^3$ the $l^{th}$ limb of the pose. An important advantage of this loss is that the model can be trained on a mix of 3D and 2D datasets simultaneously: if an image sample only has 2D annotations then $\lambda_p=0$ and $\lambda_r=0$, such that only the heatmaps are contributing to the loss.  In Section~\ref{sec:mixing2d3d} we show how having a larger corpus of 2D annotations can be leveraged to improve final 3D body pose estimates.

%
%
%

\subsection{Training Details} \label{sec:training}
The model has been trained on the entire training set for $3$ epochs, with a learning rate of $1e-3$ using batch normalization on a mini-batch of size $16$. The deconvolutional layer used to identify the heatmaps from the features computed by \textit{ResNet} has $\text{kernel size} = 3$ and $\text{stride} = 2$. The convolutional and deconvolutional layers of the encoder have $\text{kernel size} = 4$ and $\text{stride} = 2$. Finally, all the layers of the encoder use leakly ReLU as activation function with $0.2$ leakiness. The $\lambda$ weights used in the loss function were identified through grid search and set to $\lambda_{hm}=10^{-3}$, $\lambda_{p}=10^{-1}$, $\lambda_{r}=10^{-1}$ $\lambda_{\theta}=-10^{-2}$ and $\lambda_{L}=0.5$ . The model has been trained from scratch with Xavier weight initializer.

%% file: sections/experiments.tex
\begin{table*}[!t]
  \begin{center}
    \renewcommand{\arraystretch}{1.2}
    \resizebox{\textwidth}{!}{
    \setlength\tabcolsep{3.0pt}
    \begin{tabular}{lccc >{\centering}m{15mm}ccc >{\centering}m{15mm}cc}
      \toprule
      Approach & Gaming  & Gesticulating    & Greeting & Lower Stretching & Patting & Reacting & Talking & Upper Stretching & Walking  & All (mm)\\
      \midrule
      Martinez~\cite{martinez2017simple} & 109.6 & 105.4 & 119.3 & 125.8 & 93.0  & 119.7 & 111.1 & 124.5 & 130.5 & 122.1 \\
      \midrule
      \textbf{Ours --- p3d} & 138.3 & 108.5 & 100.3 & 133.3 & 117.8 & 175.6 & 93.5  & 129.0 & 131.9 & 130.4 \\
      \textbf{Ours --- p3d+rot} & 110.7 & 90.9 & 91.9 & 119.1 & 98.6 & 106.8 & 86.9 & 88.0 & 88.2 & 91.2\\
      \textbf{Ours --- p3d+hm} & 56.0 & 50.2 & 44.6 & 51.1 & 59.4  & 60.8  & 43.9  & 53.9  & 57.7  & 58.2 \\
      \textbf{Ours --- p3d+hm+rot} & 60.4 & 54.6 & 44.7 & 56.5 & 57.7 & 52.7 & 56.4 & 53.6 & 55.4 & \bf 54.7 \\
      \bottomrule
    \end{tabular}}
  \end{center}
  \caption{ Quantitative evaluation with
    Martinez~\emph{et al.}~\cite{martinez2017simple}, a state-of-the-art approach developed for front-facing cameras. Both upper and lower body reconstructions are shown as well. A comparison with our own architecture where different configurations are analyzed. Specifically, the impact of the additional branches is evaluated. Notice how the competing approach fails consistently across different actions in lower body reconstructions. This experiment emphasizes how, even a state-of-the-art 3D lifting method developed for external cameras fails on this challenging task. It also emphasizes the contribution of encoding uncertainty for achieving low-reconstruction errors. \label{tab:overall}}
\end{table*}

In the following, we thoroughly evaluate our proposed approach on our novel {\boldmath $x$}R-EgoPose dataset, we perform parameter and architecture ablations, and we evaluate on the real-world $\text{Mo}^2\text{Cap}^2$ test-set~\cite{xu2019mo2cap2} which includes 2.7K frames of real images with ground truth 3D poses of two people captured in indoor and outdoor scenes.
In addition, we show qualitative results on our controlled small-scale real-world dataset and demonstrate how our approach can be used to animate virtual characters for {\boldmath $x$}R telepresence.
Finally, we evaluate quantitatively on the Human3.6M dataset to show that our architecture generalizes well without any modifications
to the case of an external camera viewpoint.




\noindent\textbf{Evaluation protocol}: Unless otherwise mentioned, we
report the Mean Per Joint Position Error - MPJPE:
\begin{equation}
  E(\m{P}, \hat{\m{P}}) = \frac{1}{N_f}\frac{1}{N_j}\sum_{f=1}^{N_f}\sum_{j=1}^{N_j}||\m{P}_{j}^{(f)}-\hat{\m{P}}_{j}^{(f)}||_2 \label{eq:eval_protocol}
\end{equation}
where $\m{P}_j^{(f)}$ and $\hat{\m{P}}_{j}^{(f)}$ are the 3D points of the
ground truth and predicted pose at frame $f$ for joint $j$,
out of $N_f$ number of frames and $N_j$ number of joints.

To ensure high reproducibility of our results on our novel synthetic {\boldmath $x$}R-EgoPose dataset, we first evaluate our method on a randomly initialized \textit{ResNet 50}.
We intentionally do not perform any pre-training strategies given that, as we show in Sec.~\ref{sec:heatmaparch}, this affects the final results.
Our goal is to establish our {\boldmath $x$}R-EgoPose dataset as a benchmark and therefore report reproducible numbers that have been computed using a standard network architecture, trained with a simple protocol, cf.\ Sec.~\ref{sec:training}.

%
%
\subsection{Evaluation on our Egocentric Synthetic Dataset}
\label{sec:quantitative_eval}
\begin{figure*}[tb]
  \includegraphics[width=\linewidth]{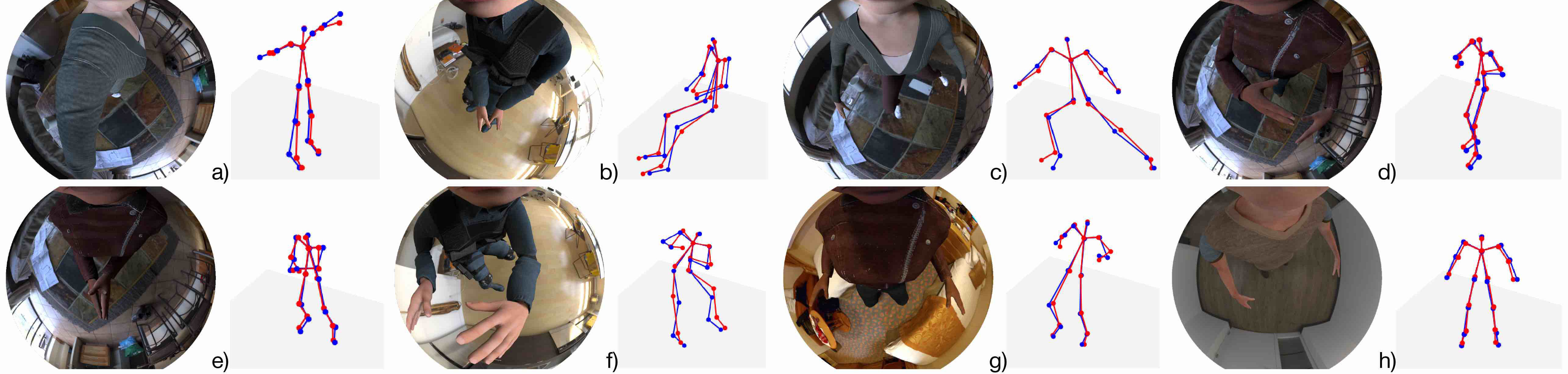}
  \caption{Qualitative results on synthetic images from our synthetic test-set. Notice that the poses are expressed with respect to the camera reference system.
  Blue poses represent ground truth, whereas poses in red correspond to predictions.\label{fig:qualitative_res}}
\end{figure*}

\noindent\textbf{Evaluation on {\boldmath $x$}R-EgoPose test-set}:
Firstly, we evaluate our approach on the test-set of our synthetic {\boldmath $x$}R-EgoPose dataset. We show qualitative results in Fig.~\ref{fig:qualitative_res}.
Unfortunately, it was not possible to establish a comparison on our dataset with state of the art monocular egocentric human pose estimation methods such as $\text{Mo}^2\text{Cap}^2$~\cite{xu2019mo2cap2} given that their code has not been made publicly available. Instead we compare with Martinez \emph{et al.}~\cite{martinez2017simple}, a recent state of the art method for a traditional external camera viewpoint. For a fair comparison, the training-set of our {\boldmath $x$}R-EgoPose dataset has been used to re-train the model of Martinez \emph{et al.}. This way we can directly compare the performance of the 2D to 3D modules.


Table~\ref{tab:overall} reports the MPJPE (Eq.~\ref{eq:eval_protocol})
for both methods showing that our approach (Ours-dual-branch) outperforms Martinez \emph{et al.}'s by 36.4\% in the upper body reconstruction, 60\% in the lower body reconstruction, and 52.3\% overall, showing a considerable improvement.

\noindent\textbf{Reconstruction errors per joint type}:
Table~\ref{tab:joints_error} reports a decomposition of the reconstruction error into different individual joint types. The highest errors are in the hands and feet. This observation is in accordance with the fact that hands and feet are often not or only barely visible. Hands can go out of the camera field of view e.g.\ by lifting or stretching the arms or may be occluded by the body. Feet are only visible when the subject looks slightly down and always cover only a very small portion of the image, due to the strong distortion. Nevertheless, our method always predicts plausible poses, even for high occlusions as displayed in Fig.~\ref{fig:animation}, Fig.~\ref{fig:qualitative_res} and Fig.~\ref{fig:qualitative_real_res}.

\noindent\textbf{Effect of the decoder branches:}
Table~\ref{tab:overall} reports an ablation study to compare the performance of three versions of our approach. We report results using: \emph{i)} only 3D pose supervision only (Ours~---~p3d); \emph{ii)} additional supervision on regressed rotations (Ours~---~p3d+rot); \emph{iii)} and on regressed heatmaps (Ours~---~ p3d+hm); finally for our novel \emph{multi-branch} auto-encoder supervised on all three signals (Ours~---~p3d+hm+rot).

The overall average error of the single branch encoder is $130.4$ mm, far from the $54.7$ mm error achieved by our novel \emph{multi-branch} architecture.
The dual branch encoders produce an error of $91.2$ mm and $58.2$ mm, respectively. Ours results clearly demonstrate that all branches contribute to our final result. Both, forcing the network to encode uncertainty of the 2D joint estimates by regressing heatmaps, as well as preserving the limb orientation information by regressing rotations, helps to estimate better 3D poses.

\begin{figure}[tb]
  \centering
  \includegraphics[width=\linewidth]{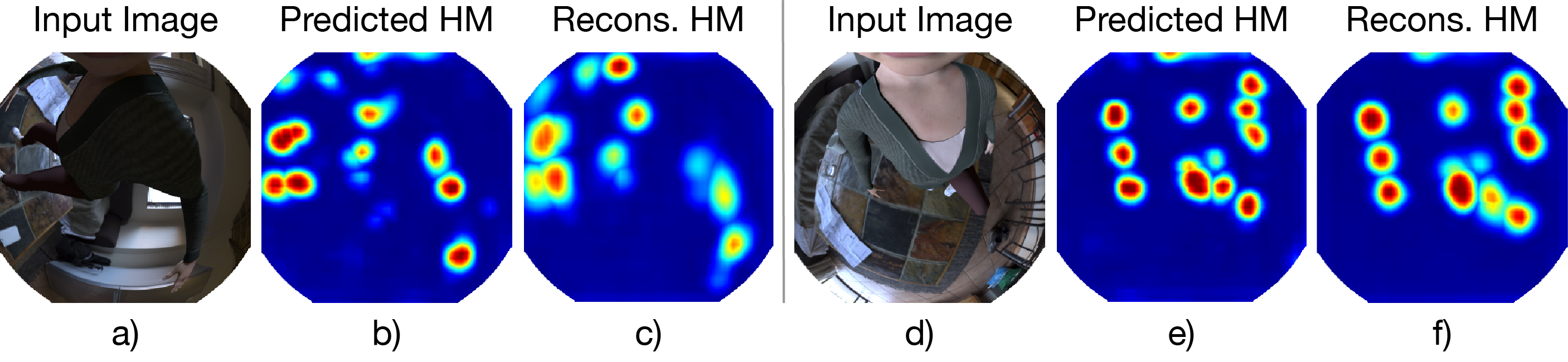}
  \caption{Reconstructed heatmaps generated by the decoder branch which can reproduce the correct uncertainty of the 2D input predictions from the pose embedding.\label{fig:hm_reconstruction}}
\end{figure}

\noindent\textbf{Encoding uncertainty in the latent space:}
Figure~\ref{fig:hm_reconstruction} demonstrates the ability of our approach to encode the uncertainty of the input 2D heatmaps in the latent vector. Examples of input 2D heatmaps and those reconstructed by the second branch of the decoder are shown for comparison.

\begin{table}[b]
  \begin{center}
    \small
    \renewcommand{\arraystretch}{1.2}
    \begin{tabular}{ll|ll}
      \textbf{Joint} & \textbf{Error (mm)} & \textbf{Joint} & \textbf{Error (mm)} \\
      \toprule
      Left Leg       & 34.33               & Right Leg      & 33.85               \\
      Left Knee      & 62.57               & Right Knee     & 61.36               \\
      Left Foot      & 70.08               & Right Foot     & 68.17               \\
      Left Toe       & 76.43               & Right Toe      & 71.94               \\
      Neck           & 6.57                & Head           & 23.20               \\
      Left Arm       & 31.36               & Right Arm      & 31.45               \\
      Left Elbow     & 60.89               & Right Elbow    & 50.13               \\
      Left Hand      & 90.43               & Right Hand     & 78.28               \\                  
    \end{tabular}
  \end{center}
  \caption{Average reconstruction error per joint using
    Eq.~\ref{eq:eval_protocol}, evaluated on the entire test-set (see Sec.~\ref{sec:dataset}) with model trained using only synthetic data.
    \label{tab:joints_error}}
\end{table}

\subsection{Character Animation Using Estimated Rotations}
The pose embedding estimation generated by the \emph{multi-branch} auto-encoder architecture contains the relevant essential information of a pose, which grants us the ability to change~/~add a representation based the a specific application. Specifically, the introduction of the rotation branch improves the overall reconstruction error, as demonstrated in Table~\ref{tab:overall}, and it is a pose definition usable for character animation.

The joint rotations estimated by the rotation-branch are expressed as local-rotations of each joint with respect to the parent node according to the skeleton hierarcy. Several rotation representations have been tested, including Euler angles, Rotation Matrices, Quaternions and the approach proposed by Zhou~\etal~\cite{zhou2019continuity}. We have not noticed any relevant improvements between Quaternions and \cite{zhou2019continuity}, however the latter demands a larger number of components per joint to express rotations.

\begin{figure*}[tb]
  \centering
  \includegraphics[width=\linewidth]{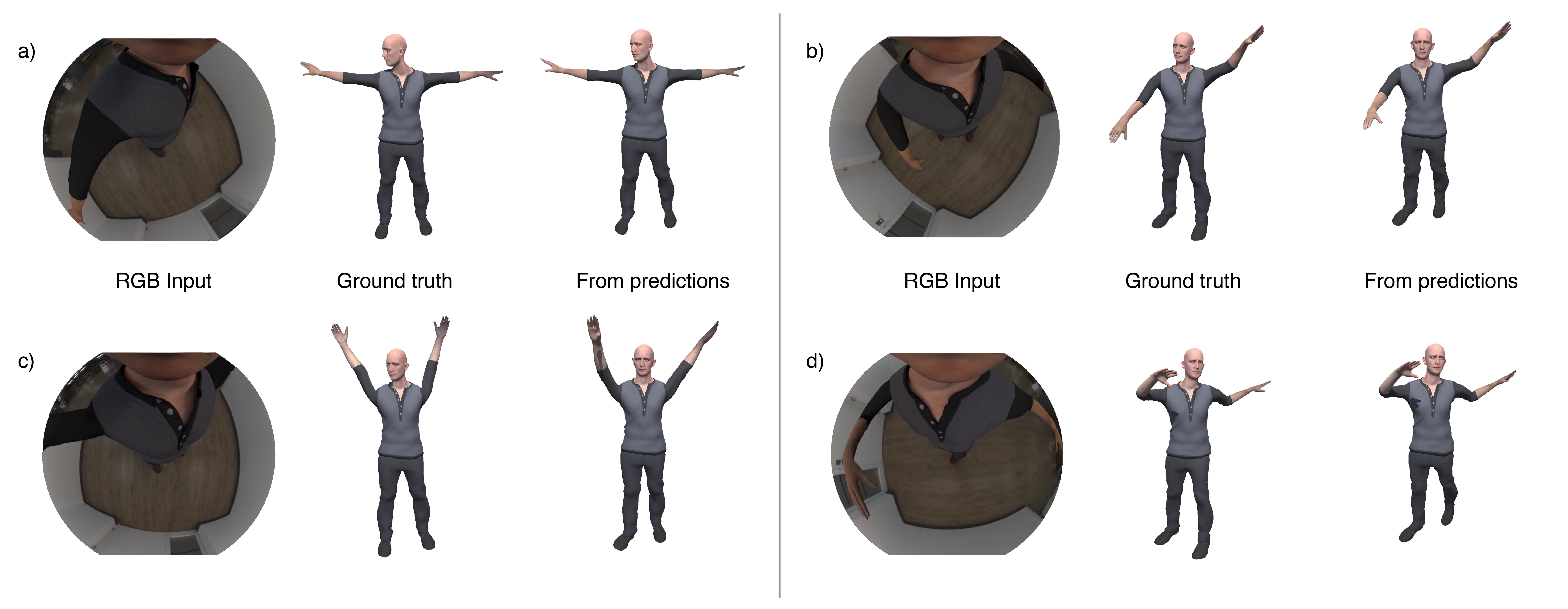}
  \caption{Character animation from the joint local rotation predictions computed from the input image. Notice how the model is able to retrieve most of the desired information even when limbs fall outside the camera field of view. \label{fig:animation}}
\end{figure*}

Example frames showing the \emph{driven character} compared against the original animation are sown in Fig.~\ref{fig:animation}. Notice how the model is able to reliably estimate the correct rotations even for poses where the avatar's limbs fall outside of the camera's field-of-view. Furthermore, there is temporal consistency between poses in consecutive frames despite estimations being computed frame-by-frame.

\begin{figure}[tb]
  \centering
  \includegraphics[width=\linewidth]{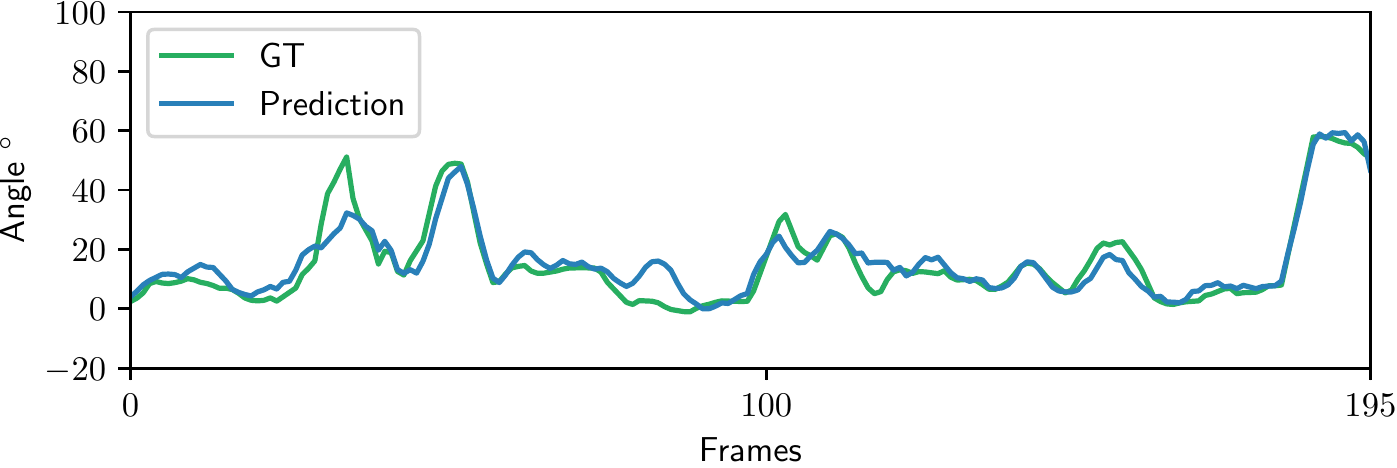}
  \caption{Analysis of the angle predictions through time for the Righ Foot in sequence of the test-set.}
  \label{fig:angle_temporal}
\end{figure}

Fig.~\ref{fig:angle_temporal} shows joint angle predictions, estimated from input images, through time. Specifically, joint angles are consistent with the ground truth. The rotations are smooth and limited ``jittering'' artefacts are introduced by the network in the predictions.

\subsection{Heatmap Estimation: Architecture Ablation} \label{sec:heatmaparch}
So far, we have used the established \textit{ResNet 50}~\cite{he2016deep} architecture in all our experiments.
In order to study the effect of the heatmap estimation network, we experiment with different architectures and initialization strategies.
Specifically, we experiment with \textit{ResNet 50}~\cite{he2016deep} and \textit{U-Net} \cite{ronneberger2015u}.
We use \textit{ResNet 50} in two variants: randomly initialized using Xavier initialization~\cite{glorot2010understanding} and pre-trained on ImageNet~\cite{russakovsky2015imagenet}.
The \textit{U-Net} is composed from a \textit{ResNet 18} backbone encoder, pre-trained on ImageNet, and a randomly initialized decoder.
The \textit{ResNet 50} consists of 24.2 million trainable parameters. The \textit{U-Net} contains 18.3 million parameters.
All variants produce the same heatmap resolution for better comparison.
The lifting networks share the same architecture and number of parameters, but have been trained specifically for each 2D pose estimation network, to accommodate its unique heatmap properties.
We additionally experimented with \textit{ResNet 101}~\cite{he2016deep},  \textit{Convolutional Pose Machines}~\cite{wei2016convolutional}, and \textit{Stacked Hourglass
Network}~\cite{newell2016stacked}. These experiments resulted in comparable performance at a higher computational cost compared to \textit{ResNet 50}, and are therefore not discussed in following.

Our experiments suggest that pre-training helps. The full pipeline using a pre-trained \textit{ResNet 50} improves the MPJPE error to $51.1$ mm, compared to $54.7$ for random initialization, see Tab.~\ref{tab:overall-2d-alternatives}.
While a recent work \cite{he2019rethinking} suggests that pre-training usually is not necessary, the authors describe two aspects where pre-training does help.
First, pre-training helps faster convergence. Second, for small datasets, pre-training helps to improve accuracy. While our synthetic dataset is large, it features less variability in scenes and subjects, compared to large real-world datasets like e.g.\ MPII \cite{andriluka142d}.

\begin{figure}[tb]
  \centering
  \includegraphics[width=\linewidth]{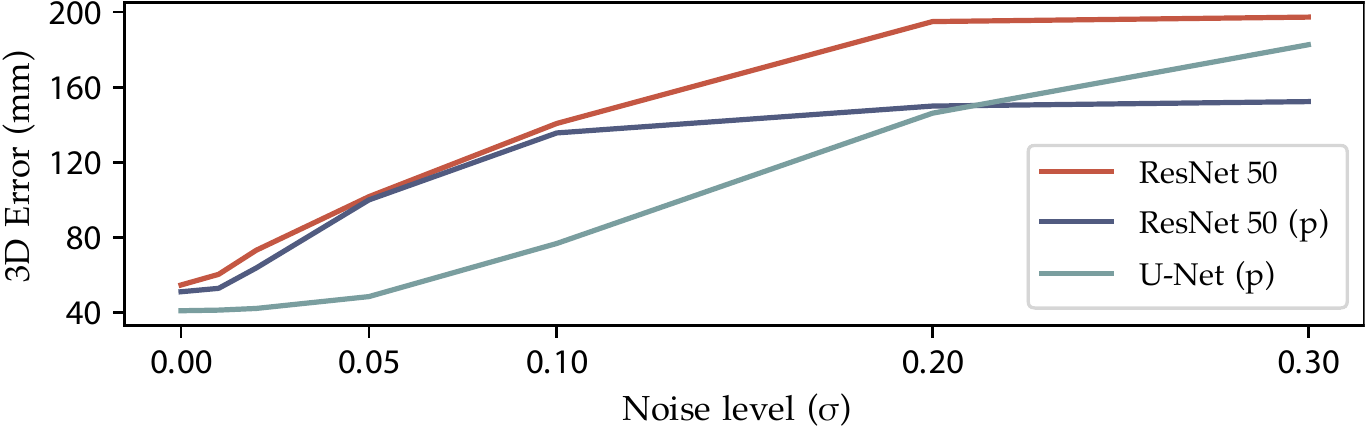}
  \caption{Performance of our proposed pipeline using different 2D pose estimation networks under the influence of white Gaussian noise in the image domain. Networks with (p) have been pretrained on ImageNet.}
  \label{fig:whitenoise}
\end{figure}

In a next step, we experiment using a \textit{U-Net} for 2D pose estimation. Using a \textit{U-Net} architecture boosts the performance of our pipeline and significantly improves the MPJPE error to $41.0$ mm.
Empirically, we found that the \textit{U-Net}-based 2D pose estimator also generalizes, to a certain extent, to real data,
predicting plausible heatmaps for  unseen data, while only having been  trained on our synthetic dataset.
The \textit{Resnet 50}-based estimator fails without prior refinement.
We hypothesize, that the improved  performance, and the observed behavior on real images, demonstrate better generalization properties of the \textit{U-Net}.
To support our hypothesis, we perform an additional experiment.
We add white Gaussian noise to the test images of our synthetic dataset and measure the performance of our pipeline using the different 2D pose estimation networks.
In Fig.~\ref{fig:whitenoise} we plot the MPJPE error under various levels of noise.
Notably, the error of the \textit{U-Net}-based pipeline increases slowly, while \textit{Resnet 50}-based pipelines produce large errors already under small noise levels.
This behavior supports our hypothesis that the \textit{U-Net} architecture features better generalization properties.

\begin{table*}[!t]
  \begin{center}
    \renewcommand{\arraystretch}{1.2}
    \resizebox{\textwidth}{!}{
    \setlength\tabcolsep{3.0pt}
    \begin{tabular}{lccc >{\centering}m{15mm}ccc >{\centering}m{15mm}cc}
      \toprule
      Configuration & Gaming  & Gesticulating    & Greeting & Lower Stretching & Patting & Reacting & Talking & Upper Stretching & Walking  & All (mm)\\
      \midrule
      ResNet 50  & 60.4 & 54.6 & \bf 44.7 & 56.5 & 57.7 & 52.7 & 56.4 & 53.6 & 55.4 & 54.7 \\
      ResNet 50  (p) & \bf 51.6 & \bf 44.6 & 64.6 & 52.4 & \bf 50.8 & \bf 44.0 & 46.5 & 51.4 & 52.8 & 51.1\\
      U-Net (p) & 52.5 & 49.2 & 72.0 & \bf 37.3 & 53.0 & 44.4 & \bf 46.1 & \bf 39.3 & \bf 37.2 & \bf 41.0\\
      \bottomrule
    \end{tabular}}
  \end{center}
  \caption{Performance analysis: different combinations of 2D pose detectors combined with the \emph{multi-branch} lifting network. All variants have been trained and tested on the synthetic dataset. Variants with (p) have been pre-trained on ImageNet. \label{tab:overall-2d-alternatives}}
\end{table*}

%

\subsection{Lifting Network:  Parameter Ablation} \label{sec:hmzsearch}
In order to validate the architecture design choices of our \emph{multi-branch} 3D pose lifting network, we perform an ablation study of two main parameters.

First, we find the optimal size of the embedding $\hat{\m{z}}$, that encodes the 3D pose, the joint rotations, and the 2D pose uncertainty.
Table \ref{tab:heatmaps_sizes} lists the MPJPE error using different sizes for $\hat{\m{z}}$ for all three different heatmap estimation networks.
Regardless of the choice of the heatmap estimation network, we find that $\hat{\m{z}} \in \mathbb{R}^{50}$ produces the best results. Smaller embeddings produce significantly higher errors, while larger embeddings only slightly impair the results.

\begin{table}[b]
  \begin{center}
    \small
    \renewcommand{\arraystretch}{1.2}
    \begin{tabular}{r|ccc}
      \multicolumn{1}{c|}{\multirow{2}{*}{\textbf{$\hat{\m{z}}$ size}}} & \multicolumn{3}{c}{\textbf{Error (mm)}} \\
       & ResNet50 & ResNet50 (p)  & UNet (p) \\
      \toprule
      10 & 70.6 & 61.0 & 45.8 \\
      20 & 67.3 & 52.5 & 45.3 \\
      50 & \textbf{54.7} & \textbf{51.1} & \textbf{41.0} \\
      70 & 55.7 & 54.5 & 41.6 \\
      100 & 58.9 & 54.2 & 41.3 \\
      500 & 61.0 & 56.0 & 41.2 \\
    \end{tabular}
  \end{center}
  \caption{Average reconstruction error per joint using
    Eq.~\ref{eq:eval_protocol}, evaluated on the entire test-set when the model architecture differs based on the size of the 
    embedding $\hat{\m{z}}$.
    \label{tab:z_sizes}}
\end{table}

Further, we study how the dimensions of the regressed heatmaps $\widetilde{\m{HM}}$ influence the results, see \ref{tab:z_sizes}.
Unsurprisingly, we find that regressing the full heatmap produces the best results.
This is in accordance with the experiments in Sec.~\ref{sec:quantitative_eval}, where we show that encoding uncertainty via regressing heatmaps helps over using them only as input.

To contribute towards fostering  fairness in Computer Vision and Machine Learning we analyze the performance of the proposed models on our diverse dataset based  on different skin tones. A comparison is shown in  Table~\ref{tab:eval_skin}.

\begin{table}[b]
  \begin{center}
    \small
    \renewcommand{\arraystretch}{1.2}
    \begin{tabular}{r|ccc}
      \multicolumn{1}{c|}{\multirow{2}{*}{\textbf{HM size}}} & \multicolumn{3}{c}{\textbf{Error (mm)}} \\
       & ResNet50 & ResNet50 (p)  & UNet (p) \\
      \toprule
      48 & \textbf{54.7} & \textbf{51.1} & \textbf{41.0} \\
      36 & 57.8 & 59.6 & 44.2 \\
      24 & 59.9 & 57.7 & 43.8 \\
      16 & 61.2 & 56.8 & 41.4 \\
       8 & 61.4 & 56.7 & 41.7 \\
    \end{tabular}
  \end{center}
  \caption{Average reconstruction error per joint using
    Eq.~\ref{eq:eval_protocol}, evaluated on the entire test-set for different
    heatmap (HM) reconstruction sizes. Notice how little uncertainty information still
    has dramatic impact on the reconstruction accuracy.
    \label{tab:heatmaps_sizes}}
\end{table}

\begin{table}[b]
  \begin{center}
    \small
    \renewcommand{\arraystretch}{1.0}
    \begin{tabular}{l|ccc}
      \multirow{2}{*}{\textbf{Skin tone}} & \multicolumn{3}{c}{\textbf{Error (mm)}} \\
      & ResNet50 & ResNet50 (p)  & UNet (p) \\
      \toprule
      White & 42.7 & 46.5 & 46.3\\
      Light European & 61.9 & 58.2 & 43.5\\
      Dark European & 63.6 & 52.0 & 35.6\\
      Dark brown & 22.5 & 28.7 & 27.5\\
      Black & 89.0 & 68.8 & 42.7\\
    \end{tabular}
  \end{center}
  \caption{Model evaluation based on skin tones.
    \label{tab:eval_skin}}
\end{table}

\subsection{Evaluation on Egocentric Real Datasets}
\label{sec:quantitative_eval}
\begin{table*}[!t]
  \begin{center}
    \footnotesize
    \setlength\tabcolsep{9.0pt}
    \renewcommand{\arraystretch}{1.2}
    \begin{tabular}{lccccccccc}
      \toprule
       INDOOR                          & walking   & sitting   & crawling  & crouching & boxing    & dancing   & stretching & waving    & total (mm) \\
      \midrule
      3DV'17~\cite{mehta2017monocular} & 48.76     & 101.22    & 118.96    & 94.93     & 57.34     & 60.96     & 111.36     & 64.50     & 76.28      \\
      VCNet~\cite{mehta2017vnect}      & 65.28     & 129.59    & 133.08    & 120.39    & 78.43     & 82.46     & 153.17     & 83.91     & 97. 85     \\
      Xu~\cite{xu2019mo2cap2}          & 38.41     & 70.94     & 94.31     & 81.90     & 48.55     & 55.19     & 99.34      & 60.92     & 61.40      \\
      {\bf Ours - ResNet 50}           & \bf 38.39 & 61.59     &   69.53   & 51.14     & \bf 37.67 & \bf 42.10 & 58.32      & \bf 44.77 & 48.16      \\
      {\bf Ours - U-Net (p)}    & 45.83     & \bf 47.24 & \bf 47.35 & \bf 45.15 & 48.72     & 47.00     & \bf 46.15  & 46.45     & \bf 46.61  \\
      \midrule
      \midrule
      OUTDOOR                          & walking   & sitting   & crawling  & crouching & boxing    & dancing   & stretching & waving    & total (mm) \\
      \midrule
      3DV'17~\cite{mehta2017monocular} & 68.67     & 114.87    & 113.23    & 118.55    & 95.29     & 72.99     & 114.48     & 72.41     & 94.46      \\
      VCNet~\cite{mehta2017vnect}      & 84.43     & 167.87    & 138.39    & 154.54    & 108.36    & 85.01     & 160.57     & 96.22     & 113.75     \\
      Xu~\cite{xu2019mo2cap2}          & 63.10     & 85.48     & 96.63     & 92.88     & 96.01     & 68.35     & 123.56     & 61.42     & 80.64      \\
      {\bf Ours - ResNet 50}           & \bf 43.60 & 85.91     & 83.06     & 69.23     & 69.32     & \bf 45.40 & 76.68      & \bf 51.38 & 60.19  \\
      {\bf Ours - U-Net (p)}   & 53.96     & \bf 52.24 & \bf 55.50 & \bf 55.65 & \bf 54.38 & 54.48     & \bf 54.46  & 56.12     & \bf 54.61  \\
      \bottomrule
    \end{tabular}
  \end{center}
  \caption{Quantitative evaluation on $\text{Mo}^2\text{Cap}^2$ dataset~\cite{xu2019mo2cap2}, both indoor and outdoor test-sets.
  Our approach outperforms all competitors by more than \textbf{21.6\%} (13.24 mm) on indoor data and more than
    \textbf{25.4\%} (20.45 mm) on outdoor data when using only the provided synthetic data for training the model. Similarly to other experiments we provide in Sec~\ref{sec:experiments}, when using a pre-trained U-Net model with the configuration defined as in Sec~\ref{sec:heatmaparch}, results improve even further: \textbf{24.9\%} (14.79 mm) and \textbf{32.28\%} (26.03 mm) respectively.
    \label{tab:comp_mo2cap2}}
\end{table*}
\noindent\textbf{Comparison with $\text{Mo}^2\text{Cap}^2$~\cite{xu2019mo2cap2}}: We
compare the results of our approach with those given by our direct
competitor, $\text{Mo}^2\text{Cap}^2$, on their real world test set including both
indoor and outdoor sequences.
For a fair comparison, we train our model solely on their provided synthetic training data (cf.\ Fig.~\ref{fig:comparison_mo2cap2}).
Table~\ref{tab:comp_mo2cap2} reports the MPJPE errors for
both methods. Our dual-branch approach substantially outperforms
$\text{Mo}^2\text{Cap}^2$~\cite{xu2019mo2cap2} in both indoor and outdoor
scenarios.
Here again, our approach using the \textit{U-Net} model pre-trained on ImageNet produces the best results.
However, indoors in a more controlled setting, both our architecture variants are almost on par.
Notice that comparison with the stereo egocentric
system EgoCap~\cite{rhodin2016egocap} on their dataset is not meaningful, due to the hugely different camera position relative to
the head (their stereo cameras are 25 cm from the head).

\begin{figure*}[tbh]
  \includegraphics[width=\linewidth]{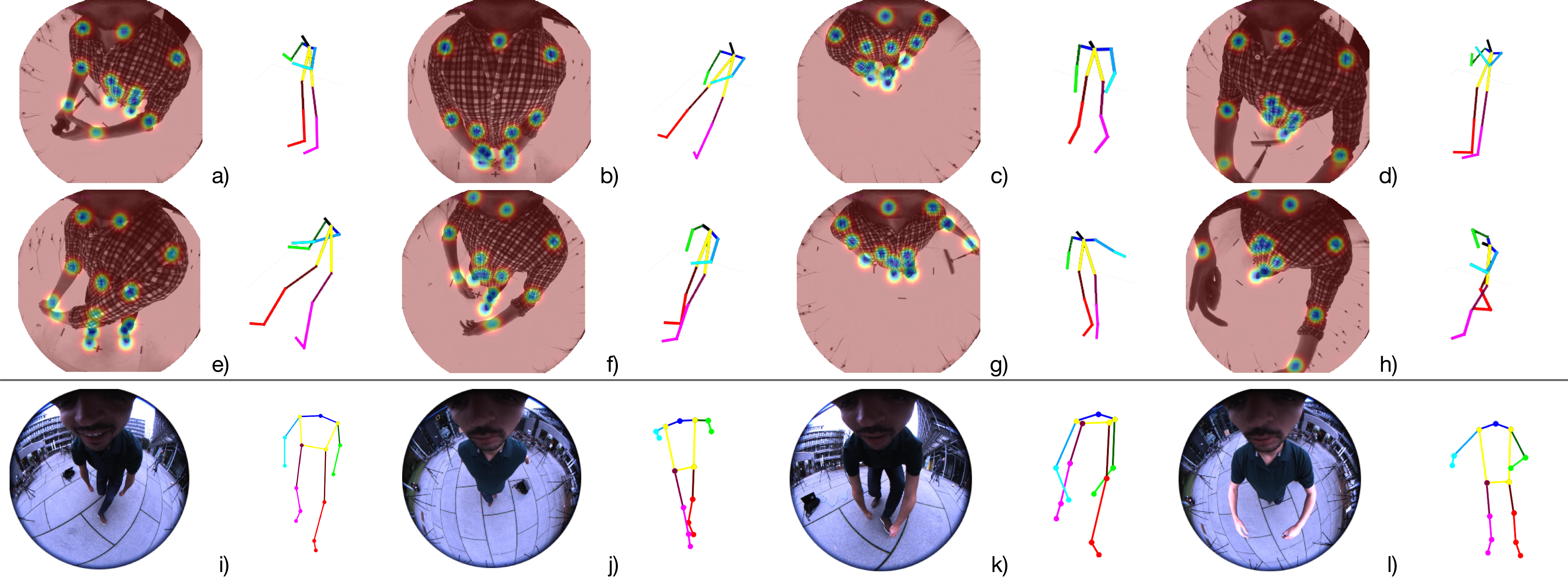}
  \caption{Qualitative results on real images captured in a studio (top) and reconstructions of
    images in the wild from $\text{Mo}^2\text{Cap}^2$~\cite{xu2019mo2cap2} (bottom).
    Notice that the poses are expressed with respect to the camera reference system.\label{fig:qualitative_real_res}}
\end{figure*}

\noindent\textbf{Evaluation on {\boldmath $x$}R-EgoPose$^\text{R}$}:
The $\sim{}10$K frames of our small scale real-world data set were
captured from a fish-eye camera mounted on a VR HMD worn by three
different actors wearing different clothes, and performing 6 different
actions. The ground truth 3D poses were acquired using a custom mocap
system. The network was trained on our synthetic corpus ({\boldmath
  $x$}R-EgoPose) and fine-tuned using the data from two of the
actors. The test set contained data from the unseen third
actor. {\boldmath $x$}R-EgoPose$^\text{R}$ is too small for meaningful numerical evaluation. However, we show qualitative examples of the input views and the reconstructed poses in Fig.~\ref{fig:qualitative_real_res}.
These results show good
generalization of the model (trained mostly on synthetic data) to real
images. 


%
\subsection{Evaluation on Front-facing Cameras}
\noindent\textbf{Comparison on Human3.6M dataset}: We show that our
proposed approach is not specific for the egocentric case, but also
provides excellent results in the more standard case of front-facing
cameras. For this evaluation, we chose the Human3.6M dataset
~\cite{ionescu2014human3, IonescuSminchisescu11}. We used two
evaluation protocols. \textit{Protocol 1} has five subjects (S1, S5,
S6, S7, S8) used in training, with subjects (S9, S11) used for
evaluation. The MPJPE error is computed on every 64th
frame. \textit{Protocol 2} contains six subjects (S1, S5, S6, S7, S8,
S9) used for training, and the evaluation is performed on every 64th
frame of Subject 11 (Procrustes aligned MPJPE is used for evaluation).
The results are shown in Table~\ref{tab:h36m} from where it can be
seen that our approach is on par with state-of-the-art methods,
scoring second overall within the non-temporal methods.

\begin{table*}[!t]
  \begin{center}
    \resizebox{\textwidth}{!}{
    \setlength\tabcolsep{3.0pt}
    \renewcommand{\arraystretch}{1.2}
    \begin{tabular}{l|cccccccccccccc}
      \toprule
      \textbf{Protocol \#1} & Chen                           & Hossain                       & Dabral                      & Tome                        & Moreno                      & Kanazawa               & Zhou 
                            & Jahangiri                      & Mehta                         & Martinez                    & Fang                        & Sun                         & Sun                    & \bf Ours \\
                            & \cite{chen20173d}              & \cite{hossain2018exploiting}* & \cite{dabral2017structure}* & \cite{tome2017lifting}      & \cite{moreno20173D}         & \cite{kanazawa2018end} & \cite{zhou2018monocap}
                            & \cite{jahangiri2017generating} & \cite{mehta2017monocular}     & \cite{martinez2017simple}   & \cite{fang2018learning}     & \cite{sun2017compositional} & \cite{sun2018integral} &          \\ 
      \midrule
      \textbf{Errors (mm)}  & 114.2                          & 51.9                          & 52.1                        & 88.4                        & 87.3                        & 88.0                   & 79.9
                            & 77.6                           & 72.9                          & 62.9                        & 60.4                        & 59.1                        & \bf 49.6               & 51.3     \\
      \midrule
      \midrule
      \textbf{Protocol \#2} & Yasin                          & Hossain                       & Dabral                      & Rogez                       & Chen                        & Moreno                 & Tome
                            & Zhou                           & Martinez                      & Kanazawa                    & Sun                         & Fang                        & Sun                    & \bf Ours \\
                            & \cite{yasin2016dual}           & \cite{hossain2018exploiting}* & \cite{dabral2017structure}* & \cite{rogez2016mocap}       & \cite{chen20173d}           & \cite{moreno20173D}    & \cite{tome2017lifting}
                            & \cite{zhou2018monocap}         & \cite{martinez2017simple}     & \cite{kanazawa2018end}      & \cite{sun2017compositional} & \cite{fang2018learning}     & \cite{sun2018integral}            \\
      \midrule
      \textbf{Errors (mm)}  & 108.3                          & 42.0                          & 36.3                        & 88.1                        & 82.7                        & 76.5                   & 70.7
                            & 55.3                           & 47.7                          & 58.8                        & 48.3                        & 45.7                        & \bf 40.6               & 42.3    \\      
    \end{tabular}}
  \end{center}
  \caption{Comparison with other state-of-the-art approaches on the Human3.6M dataset (front-facing cameras). Approaches with * make use of temporal information. No specific modifications have been applied to our architecture: UNet 2D pose detector pre-trained on ImageNet has been used to estimate joint-heatmaps fed through our dual-branch auto-encoder architecture, since rotation information is not available for these data. \label{tab:h36m}}
\end{table*}

%
%


%
%

%
%
%

%

\subsection{Mixing 2D and 3D Ground Truth Datasets} \label{sec:mixing2d3d}
An important advantage of our architecture is that the model can be
trained on a mix of 3D and 2D datasets simultaneously: if an image
sample only has 2D annotations but no 3D ground truth labels, the
sample can still be used, only the heatmaps will contribute to the
loss. We evaluated the effect of adding additional images with 2D but
no 3D labels on both scenarios: egocentric and front-facing
cameras. In the egocentric case we created two subsets of the
{\boldmath $x$}R-EgoPose test-set. The first subset contained $50$\%
of all the available image samples with both 3D and 2D labels. The
second contained $100$\% of the image samples with 2D labels, but only
50\% of the 3D labels. Effectively the second subset contained twice
the number of images with 2D annotations only.
Table~\ref{tab:quantitative:labels:ego_hmd} compares the results
between the subsets, where it can be seen that the final 3D pose
estimate benefits from additional 2D annotations. Equivalent behavior
is seen on the Human3.6M dataset. Table
\ref{tab:quantitative:labels:h36m} shows the improvements in
reconstruction error when additional 2D annotations from COCO
\cite{lin2014microsoft} and MPII \cite{andriluka142d} are used.

\begin{table}[!b]
  \centering
  \small
  \renewcommand{\arraystretch}{1.2}
  \resizebox{0.35\linewidth}{!}{
  \begin{subtable}[t]{0.4\linewidth}
    \centering
    \begin{tabular}{lll}
      3D & 2D           & Error (mm) \\
      \toprule
      50$\%$ & 50$\%$      & 68.04  \\
      50$\%$ & 100$\%$    & 63.98  \\
    \end{tabular}
    \caption{{\boldmath $x$}R-EgoPose}\label{tab:quantitative:labels:ego_hmd}
  \end{subtable}}
  \hfill
  \resizebox{0.55\linewidth}{!}{
  \begin{subtable}[t]{0.6\linewidth}
    \begin{tabular}{ll}
      Training dataset & Error (mm) \\
      \toprule
      H36M                & 67.9       \\
      H36M + COCO + MPII  & 53.4       \\
    \end{tabular}
    \caption{Human3.6M}\label{tab:quantitative:labels:h36m}
  \end{subtable}}
  \caption{Having a larger corpus of 2D annotations can be leveraged
    to improve final 3D pose
    estimation \label{tab:quantitative:labels}}
\end{table}

%% file: sections/conclusion.tex
We have presented a solution to the problem of 3D body pose estimation from a monocular camera installed on a HMD. Given a single image, our fully differentiable network estimates heatmaps and uses them as an intermediate representation to regress 3D poses using a novel \emph{multi-branch} auto-encoder.
This new architecture design was fundamental for accurate reconstructions in our challenging dataset, with over $24\%$ accuracy improvement on competitor datasets and that proves to generalize to the more generic 3D human pose estimation from front-facing cameras task with state-of-the-art performance.
We have shown how the proposed architecture can be used to drive a virtual avatar directly from the estimations of the network, a fundamental step towards telepresence in virtual or augmented reality.

Finally, we have also introduced the {\boldmath $x$}R-EgoPose dataset, a new large scale photo-realistic synthetic dataset that was essential for training and will be made publicly available to promote research in this exciting area. While our results are state-of-the-art, there are a few failures cases due to extreme occlusion and the inability of the system to measure hands when they are out of the field of view. Adding additional cameras to cover more field of view and enable multi-view sensing is the focus of our future work.